\documentclass[titlepage]{article}

\usepackage[square,numbers]{natbib}

\usepackage{arxiv}

\usepackage[utf8]{inputenc} % allow utf-8 input
\usepackage[T1]{fontenc}    % use 8-bit T1 fonts
\usepackage{hyperref}       % hyperlinks
\usepackage{url}            % simple URL typesetting
\usepackage{booktabs}       % professional-quality tables
\usepackage{amsfonts}       % blackboard math symbols
\usepackage{nicefrac}       % compact symbols for 1/2, etc.
\usepackage{microtype}      % microtypography
\usepackage{lipsum}		% Can be removed after putting your text content
\usepackage{graphicx}

\usepackage{doi}
\usepackage[english]{babel}

\title{Interpreting Audiograms with Multi-stage Neural Networks}

\author{
    Shufan Li\\
	Orka Labs Inc.\\
	\texttt{lishufan@hiorka.com}
	\And
	Congxi Lu\\
	Orka Labs Inc.\\
	\texttt{chauncey@hiorka.com} \\
	\And
	Linkai Li\\
	Orka Labs Inc.\\
	\texttt{linkai@hiorka.com} \\
	\AND
	Jirong Duan\\
	ENT\&Audiology Center\\
	Xinhua Hospital\thanks{Xinhua Hospital is affiliated to School of Medicine, Shanghai Jiao Tong University }\\
	\&Punan Hospital\\
	\texttt{jirong\_duan@163.com}
	\And
	Xinping Fu\thanks{Xinping Fu is studying at Ear Institute, University College London} \\
	Ear Institute\\
	University College London\\
	\texttt{xinping\_fu@163.com}
	\And
	Haoshuai Zhou\thanks{Corresponding Author}\\
	Orka Labs Inc.\\
	\texttt{haoshuai@hiorka.com}
}

\begin{document}

\maketitle

\renewcommand{\thefootnote}{\arabic{footnote}}

\begin{abstract}
	 Audiograms are a particular type of line charts representing individuals' hearing level at various frequencies. They are used by audiologists to diagnose hearing loss, and further select and tune appropriate hearing aids for customers. There have been several projects such as \textit{Autoaudio} that aim to accelerate this process through means of machine learning.  But all existing models at their best can only detect audiograms in images and classify them into general categories. They are unable to extract hearing level information from detected audiograms by interpreting the marks, axis, and lines. To address this issue, we propose a \textit{Multi-stage Audiogram Interpretation Network} (\textit{MAIN}) that directly reads hearing level data from photos of audiograms.  We also established \textit{Open Audiogram}, an open dataset of audiogram images with annotations of marks and axes on which we trained and evaluated our proposed model. Experiments show that our model is feasible and reliable.
\end{abstract}

% keywords can be removed

\section{Introduction}
Hearing loss is becoming an increasingly widespread health problem \cite{world2018addressing, gbd2018global, orji2020global}. WHO estimates that around 2.5 billion people, or 1/4 of the global population will experience hearing loss by 2050 \cite{chadha2021world}.  Fortunately, hearing aids can compensate for hearing deficits and considerably improve individuals’ quality of life \cite{chisolm2007systematic, shield2006evaluation, dillon2008hearing}.  Before hearing aids can be delivered to a customer, appropriate adjustments to the device (known as the fitting process) must be performed according to the audiogram of the customer. 

In response to the growing demand, the U.S Senate passed the Over-the-Counter Hearing Aid Act of 2017 which created a class of hearing aids directly available to consumers without the involvements of licensed professionals \cite{library_of_congress}. To make these products viable choices for consumers, automated fitting processes are of necessity. While some of these products can measure hearing loss and adjust the device accordingly (self-fitting), it is more desirable to use results from specialized hearing tests when a user is already in possession of an audiogram.

Outside the U.S. many parts of the world are experiencing a shortage of audiologists which limited access to well-tuned hearing aids \cite{kamenov2021ear}. Moreover, even for professional audiologists, the task of interpreting audiograms can be laborious and prone to human error. Hence, an effective and reliable automatic audiogram interrupter can be a game-changer as it allows a limited supply of audiologists to serve a larger population and improves the experiences of OTC hearing aids' customers.

While there have been attempts to directly classify the types of hearing loss (if any) from audiogram images \cite{Charih2020DataDrivenAC, lee2010using, Crowson2020AutoAudioDL}, they are only able to provide class-based qualitative descriptions of audiograms. They cannot recover the full information, specifically the precise hearing level at different frequencies, which are crucial for tuning hearing aids. Hence, these models are not particularly helpful in the process of fine-tuning a hearing aid, and alternative methods which can extract detailed hearing level information would be more desirable.

To address this gap between growing demands and the limited capacity of existing models, in this paper we tackle the task of audiogram interpretation, which we define as recovering the $(frequency,hearing\_level)$ tuple from an arbitrary audiogram photo. Compared with audiogram classification, audiogram interpretation requires extracting the complete information necessary to reconstruct a digital copy of the original audiogram, and is therefore more challenging.  We introduce a new dataset Open Audiogram consisting of 420 photos of 34 audiograms taken in real-world scenarios with varying lighting, angle, and backgrounds. In addition, we propose a baseline Multi-stage Audiogram Interpretation Network. In the first stage, we detect audiograms in an arbitrary image and crop each audiogram by its bounding box to remove potential interference from the background. In the second stage, we attempt to rectify the distortion caused by the camera angle by applying a perspective transform. In the third stage, we detect marks and axis labels from cropped, rectified images. We use axis labels to reconstruct a coordinate system, and then project mark coordinates onto the fitted axes to recover $(frequency, hearing\_level)$ tuples. RANSAC algorithm is adopted in this process to remove outliers caused by misclassification. An overview of this pipeline is shown in Figure \ref{fig:fig2}. Experiments show that our proposed baseline is feasible with a test accuracy of 84\%. Further, 94\% of the detected tuples are within a margin of $\pm5$ dB HL in hearing level, which is the smallest possible error in hearing level since all hearing level entries are usually multiples of 5 (Table \ref{table:table3}).

\section{Background}
At a high level, audiograms are visual representations of an individual's hearing level. The task of audiogram interpretation is the task of retrieving quantitative hearing level information from images of audiograms. In this section, we give formal definitions of this task and the relevant concepts. Additionally, we summarize prior related works.

\subsection{Audiogram Interpretation}
\begin{figure}
	\centering\
	   \includegraphics[width=0.35\linewidth]{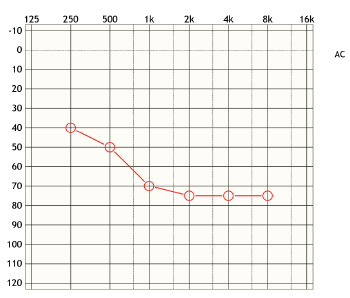}

% 	\fbox{\rule[-.5cm]{4cm}{4cm} \rule[-.5cm]{4cm}{0cm}}
	\caption{A typical example of audiograms. The X-Axis is frequency in Hz. The Y-Axis is hearing level in dB HL. The line and its marks represent an individual's hearing level at different frequencies.}
	\label{fig:fig1}
\end{figure}

An \textit{audiogram} is a line chart representing a person's hearing level. As is shown in Figure \ref{fig:fig1}, a typical audiogram consists of the following visual elements: X-axis representing frequencies ranging from 125 Hz to 16000 Hz, Y-axis representing hearing levels in dB HL ranging from -10 dB HL to 120 dB HL, and marks whose coordinates indicate an individual's hearing level at different frequencies. The X-axis has log-scale tick marks at 125 Hz, 250 Hz, ..., 8000 Hz, 16000 Hz. The Y-axis has linear, uniform tick marks with a step size of 10 dB HL. 

Compared with a general line chart whose marks have an infinite, continuous sample space contained in  $\mathbb {R}^2$, audiograms are easier to interpret because we can utilize this characteristic and map the fitted real-valued coordinates to its nearest neighbor in the sample space. Another important characteristic we make use of is the fact that X tick marks and Y tick marks have no overlapping values. So if we can recognize the number of a tick mark, we can automatically determine which axis it belongs to. This reduces the difficulty of inferring the direction of axes.

The \textit{digital representation of an audiogram} is the set $G=\{(x_i, y_i)\}|i=1, 2, ..., n\}$ containing the coordinates of all the marks observed in an audiogram, where $x_i\in \{125, 250, ..., 8000, 16000 \}$ are frequencies in Hz and $y_i\in \{-10, 0, ..., 110, 120 \}$ are hearing levels in dB HL. Let $\Omega$ be the set of all audiogram images and $\Gamma$ be the set of all possible digital representations of audiograms. There exists a natural mapping $F:\Omega \rightarrow \Gamma$ which maps every audiogram image to its actual digital representation. We define the task of audiogram interpretation as the task of finding a computable, accurate approximation mapping $f:\Omega \rightarrow \Gamma$ of $F$ such that for an image $A\in \Omega$ we can use $f(A)$ to find an approximation of $F(A)$. This paper is the first to study the automatic interpretation of audiograms.

\subsection{Related Works}
% \lipsum[5]
% \begin{equation}
% 	\xi _{ij}(t)=P(x_{t}=i,x_{t+1}=j|y,v,w;\theta)= {\frac {\alpha _{i}(t)a^{w_t}_{ij}\beta _{j}(t+1)b^{v_{t+1}}_{j}(y_{t+1})}{\sum _{i=1}^{N} \sum _{j=1}^{N} \alpha _{i}(t)a^{w_t}_{ij}\beta _{j}(t+1)b^{v_{t+1}}_{j}(y_{t+1})}}
% \end{equation}
The related works of audiogram interpretation mainly fall into two categories: (1) audiogram classification and (2) chart data extraction.
\subsubsection{Audiogram classification}
In contrast to audiogram interpretation, the task of audiogram classification only attempts to retrieve a qualitative summary of audiograms based on classes. There have been a variety of works in this field, most of which rely on expert-tuned rules \cite{margolis2011audiogram, carhart1945improved, guild1932method}. In recent years, some methods utilizing machine learning techniques to distinguish different audiograms have been proposed. Charih et al. \cite{Charih2020DataDrivenAC} make use of decision forest to classify audiograms based on configuration, symmetry, and severity.  Lee et al. \cite{lee2010using} leverage K-means clustering to categorize audiogram shapes into six basic types and five subtypes.   Crowson et al. \cite{Crowson2020AutoAudioDL} employ ResNet-101 to differentiate audiograms of individuals with conductive, sensorineural, mixed, or no hearing loss. Nonetheless, all these methods only give a rough summary of certain properties of audiograms. They cannot provide enough details for a hearing aid prescription.

\subsubsection{Chart Data Extraction}
Chart Data Extraction refers to the general task of retrieving quantitative information for chart images. Existing Chart Data Extraction algorithms fall into two categories:  automatic and semiautomatic.  Semiautomatic algorithms typically involve extensive, time-consuming user interactions.   For example, WebPlot-Digitizer \cite{Rohatgi2020}, a famous semiautomatic digitizing tool, requires users to manually label and calibrate a lot of data points. ChartSense \cite{Jung2017ChartSenseID}, another interactive data extraction system, also requires the user to manually identify chart areas and intervals of data points.  In contrast, automatic algorithms require no manual work. One common approach is to combine image processing algorithms with  OCR.  Histogram-based techniques and spatial filtering are generally used to locate coordinate axes and data points,  while OCR is employed to detect textblocks. Some recent works replace these traditional image processing techniques with deep learning models. Scatteract \cite{Cliche2017ScatteractAE}, for example, use ReInspect to detect data points,  tick marks and axis labels.  However,these methods are only tested on high-quality screenshots where there is little noise, occlusion, or rotation in the image, which are drastically different from photos in our dataset. Additionally, they did not make use of the aforementioned limited sample space of mark coordinates and tick marks. Hence they are at best sub-optimal solutions to our problem.

Our work lies at the intersection of Audiogram Classification and Chart Data Extraction.  To the best of our knowledge, we are the first to propose a feasible model which automatically extracts the complete audiogram information from real-world photos under diverse lighting, occlusion, and rotation.

\begin{figure}
	\centering\
	   	   \includegraphics[width=0.9\linewidth]{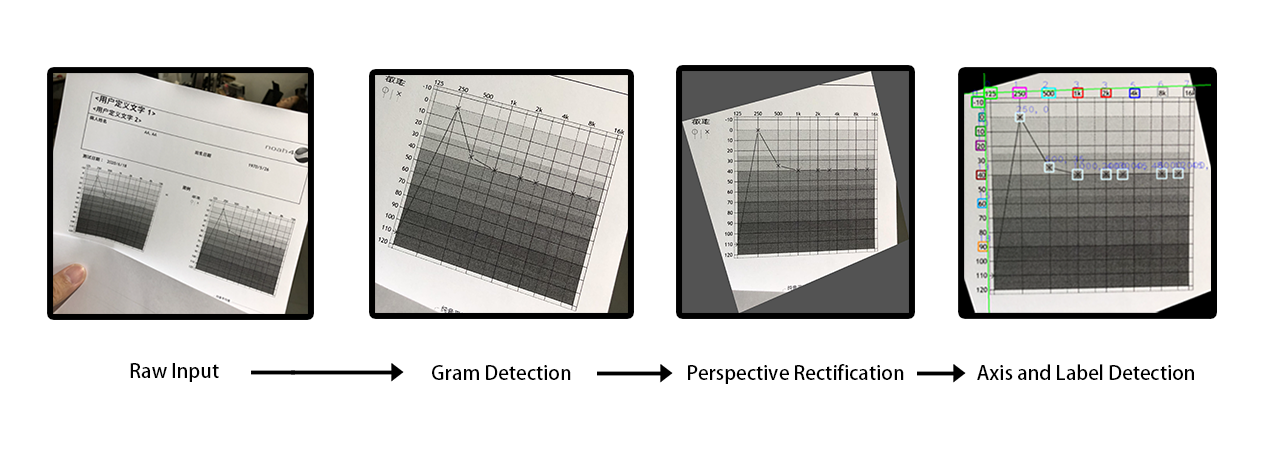}
% 	\fbox{\rule[-.5cm]{4cm}{4cm} \rule[-.5cm]{4cm}{0cm}}
	\caption{MAIN's overall structure. It consists of three stages: (1) gram detection, (2) perspective rectification (3) axis and label detection.}
	\label{fig:fig2}
\end{figure}

\section{The Open Audiogram Dataset}
\begin{figure}
	\centering\
	   \includegraphics[width=0.5\linewidth]{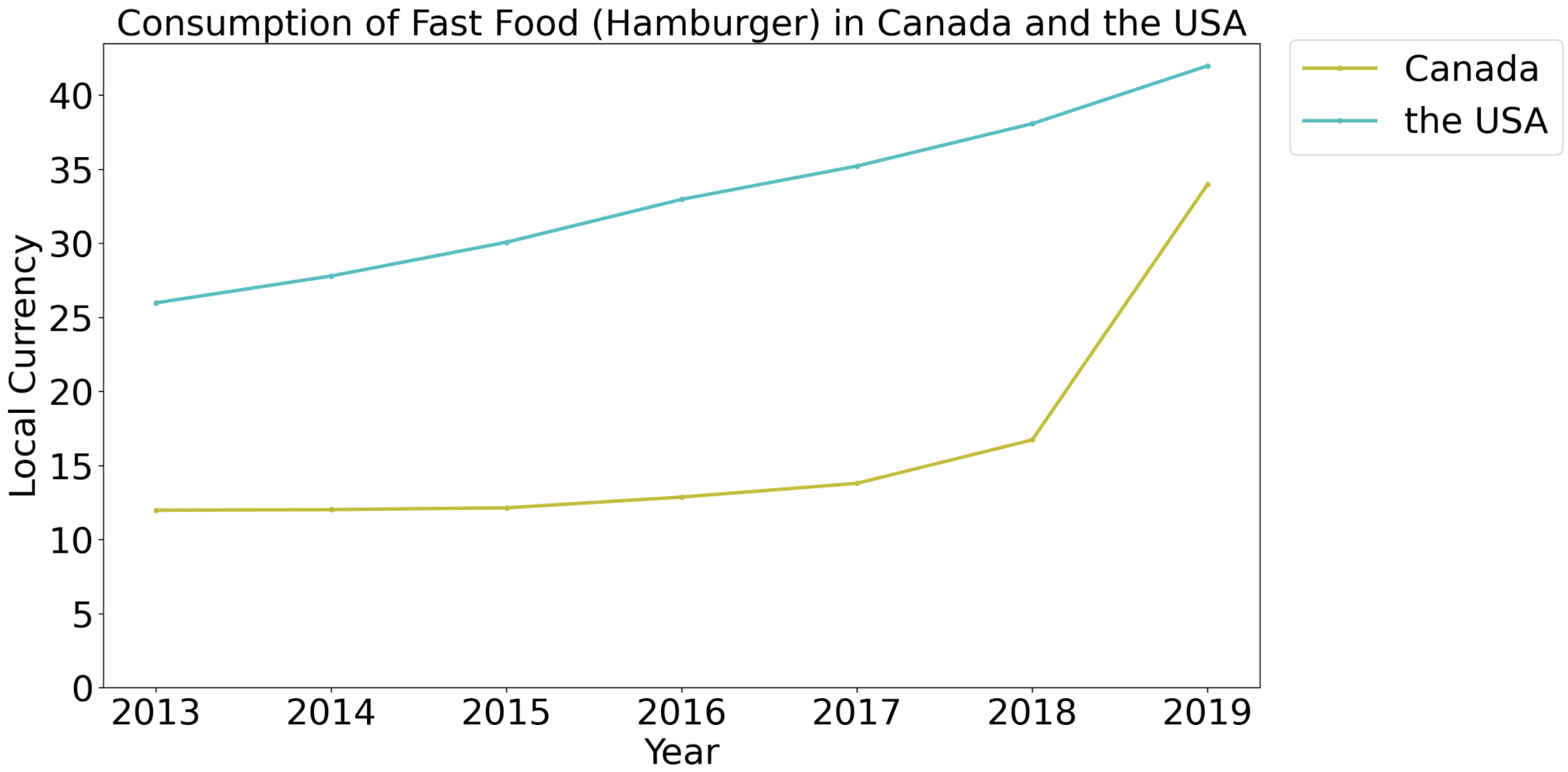}

% 	\fbox{\rule[-.5cm]{4cm}{4cm} \rule[-.5cm]{4cm}{0cm}}
	\caption{An example from AutoChart dataset, which is composed of mainly synthetic chart images.}
	\label{fig:figauto}
\end{figure}

In this paper,  we present a challenging dataset for audiogram recognition. Most of the existing chart recognition datasets such as AutoChart consist of synthetic chart images with almost no variation in lightening conditions and perspective angles (Figure \ref{fig:figauto}). These images also do not follow the aforementioned characteristics of audiograms. Hence, they are unfit for audiogram recognition. To address this apparent gap, we created Open Audiogram, a dataset consisting of 420 camera photos of 34 audiograms and additional 30 scanned images of different audiograms. 

\subsection{Image Preparation}
\begin{figure}
	\centering\
	   \includegraphics[width=0.8\linewidth]{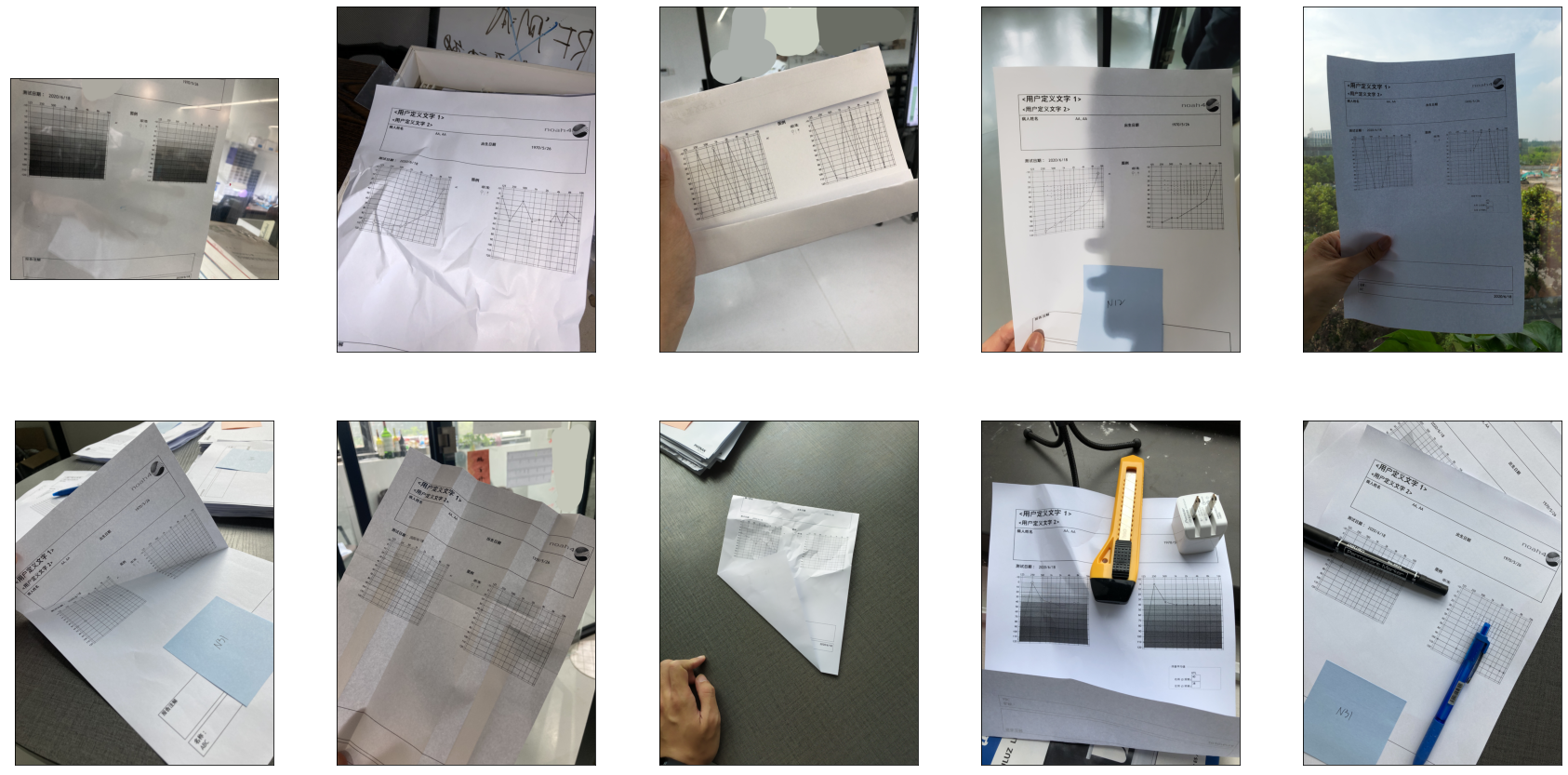}

% 	\fbox{\rule[-.5cm]{4cm}{4cm} \rule[-.5cm]{4cm}{0cm}}
	\caption{Examples of photos in our Open Audiogram dataset taken under varying lighting conditions, occlusions and camera angles. }
	\label{fig:fig3}
\end{figure}

\begin{figure}
	\centering\
	   \includegraphics[width=0.25\linewidth]{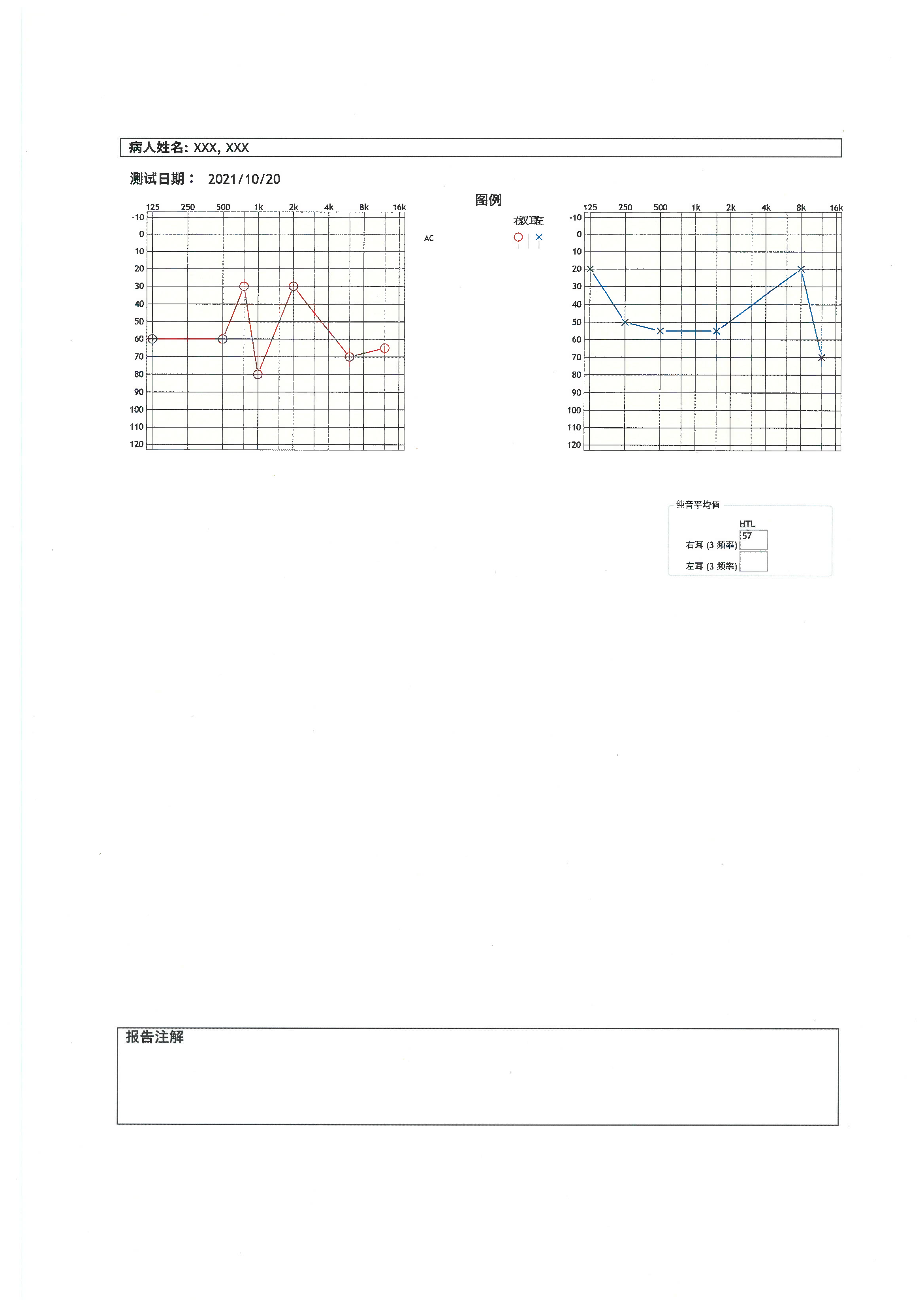}

% 	\fbox{\rule[-.5cm]{4cm}{4cm} \rule[-.5cm]{4cm}{0cm}}
	\caption{An Example of our scanned audiograms}
	\label{fig:figscanned}
\end{figure}

We first collected raw audiogram images generated by Noah 4 System \cite{noah4} from our audiologists' patient records and filterd out all sensitive data. Then these audiograms were printed on standard A4-sized paper. We took a total of 420 images of these printed audiograms under various angles between 0 and 45 degrees (measured by the angle of the camera direction and the normal vector of the paper).   We also intentionally took the photos at different locations in an office building to adjust the balance between artificial indoor lighting and natural skylight coming from the window. In some photos, there are cast shadows of other objects on the paper with the edge of the shadow crossing the audiogram region. In some photos, obstructions such as pens are purposely placed on the audiogram region. In some photos, the paper is bent or folded (Figure \ref{fig:fig3}). During this process, all sensitive information such as human faces and computer screens are masked out for privacy concerns. These intentional occlusions and distortions make the dataset particularly challenging. To evaluate the impact of these distortions, we generated an additional set of 30 audiograms, printed them, and scanned them using a laser scanner. These undistorted images look nearly identical to the raw image (Figure \ref{fig:figscanned}).


\subsection{Annotation}
 There are fours levels of annotations.
\begin{itemize}
\item Level 1: With the raw $(frequency,hearing\_level)$ pairs used to generate the audiogram
\item Level 2: With the bounding boxes of the gram region
\item Level 3: With the bounding polygons (mostly quadrilateral with few exceptions) of the chart region
\item Level 4: With the bounding boxes of all marks and axis present in the gram region

\end{itemize}

The level 1 annotations alone provide sufficient information to evaluate the performance of any audiogram interpretation algorithm because it allows the comparison between the predicted $(frequency,hearing\_level)$ pairs and ground truths. Due to the multistage nature of our proposed baseline, it is not very helpful to annotate all images at all levels. For example, the marks and axis detection is performed on cropped images, and hence providing level 4 annotation on raw photos is meaningless. Therefore, we have varying numbers of images at each annotation level, which are summarized in Table \ref{table:table2}.

To our best knowledge, our dataset is the first audiogram dataset focused on data extraction in real-world scenarios. It can serve as a benchmark for evaluating different existing approaches and facilitate future works on audiogram interpretation.

\begin{table}
	\caption{Audiogram Annotations}
	\centering
	\begin{tabular}{cccc}
		\toprule
		 & \multicolumn{2}{c}{Photo}    &   Scanned             \\
		\cmidrule(r){1-4}
	 	 & Cropped     & Raw     & Raw  \\
		\midrule
		Total      & 410    & 420  & 30   \\
		level1     & 0      & 420  & 30    \\
		level2     & 0      & 420 & 0 \\
		level3     & 410    & 0  & 0\\
		level4     & 223    & 0   & 0\\
		\bottomrule
	\end{tabular}
	\label{table:table2}
\end{table}
\section{Model Architecture}

Our proposed Multi-stage Audiogram Interpretation Network (MAIN) consists of three stages: gram detection, perspective rectification, and axis and label detection (Figure \ref{fig:fig2}). In the first stage, we use a Faster-RCNN to detect the gram region containing the axis and marks.  In the second stage, we leverage a Mask-RCNN in combination with traditional image processing techniques such as hough line transform to get an estimate of vanishing points and apply a perspective transform to rectify the image. In the last stage, we employ another Faster-RCNN to detect axis labels and marks. We use RANSAC to fit two axes from the label coordinates, and then project mark coordinates onto the fitted axes to obtain their frequencies and hearing levels.
\subsection{Gram Detection}
The task of the gram detection network is to identify the bounding box of the gram areas containing axis labels and marks. We used a Faster-RCNN with ResNet and FPN backbone for this task. Existing benchmarks show that such architecture has a satisfying performance on the Common Objects in Context (COCO) Dataset and is therefore suitable for this task \cite{wu2019detectron2}. 

\subsection{Perspective Rectification}
\begin{figure}
	\centering\
	   \includegraphics[width=0.35\linewidth]{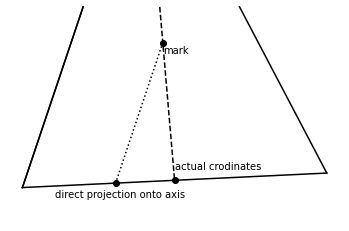}

% 	\fbox{\rule[-.5cm]{4cm}{4cm} \rule[-.5cm]{4cm}{0cm}}
	\caption{An example of perspective distortion, in which the naive projection deviates from the ground truth because parallel lines are no longer parallel .}
	\label{fig:fig5}
\end{figure}

Since the final $(frequency, hearing\_level)$ predictions are generated by projecting detected data points on the detected axes, it is crucial that the image does not contain a vanishing point. Parallel lines in the audiogram must stay parallel in the image. For example, in Figure \ref{fig:fig5}, the projected coordinate of the mark deviates from the ground truth by a huge margin because of the perspective distortion. We propose two methods to rectify perspective distortions caused by camera angles. The first primarily relies on line detection to get an estimation of vanishing points, the second utilizes Mask-RCNN to predict the four corners of the gram region. 
\begin{figure}
	\centering\
	   \includegraphics[width=0.9\linewidth]{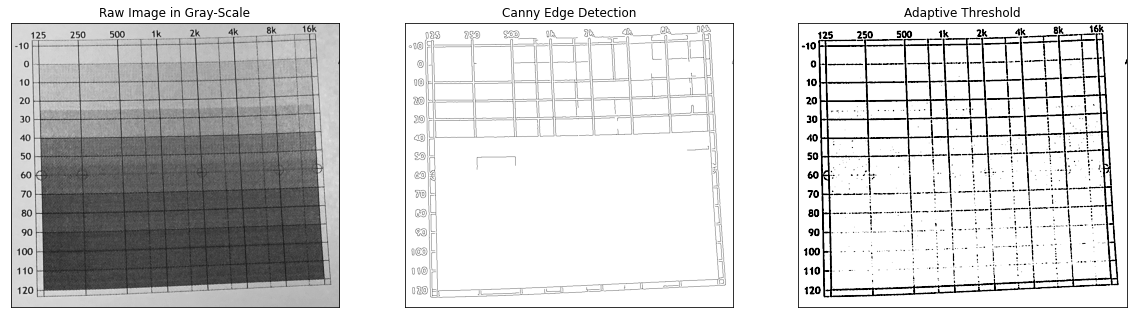}

% 	\fbox{\rule[-.5cm]{4cm}{4cm} \rule[-.5cm]{4cm}{0cm}}
	\caption{A comparison of line extraction methods. Our adaptive threshold method is able to better preserve gridline information. }
	\label{fig:fig6}
\end{figure}
\begin{figure}
	\centering\
	   \includegraphics[width=1.0\linewidth]{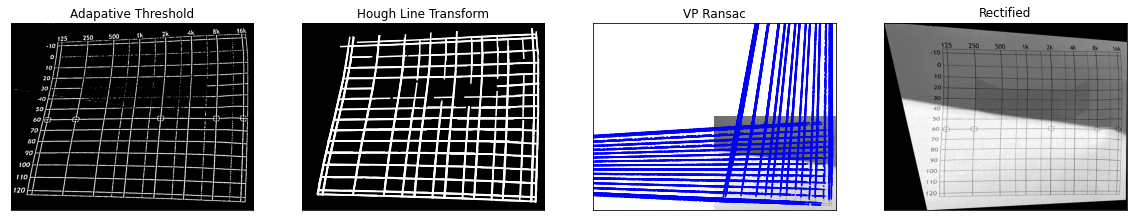}

% 	\fbox{\rule[-.5cm]{4cm}{4cm} \rule[-.5cm]{4cm}{0cm}}
	\caption{An illustration of perspective rectification method in 4.2.1. It has four stages. (1) Binarize the image by adaptive threshold. (2) Extract long lines using hough line transform. (3) Fit two vanishing points using RANSAC. (4) Apply an inverse perspective transform. }
	\label{fig:figrec}
\end{figure}

\subsubsection{Line-Detection-Based Approach}
Chaudhury et al. proposed a method to estimate vanishing points by applying RANSAC on extracted edgelets. They extract edgelets from images by applying a "Haris Corner Detector style" filter on the image and compute the eigenvalues and eigenvectors of the Gaussian weighted covariance matrix in a square neighborhood \cite{chaudhury2014auto}. A later work modified this approach by using canny edge detection in combination with probabilistic hough line transform to detect edgelets instead \cite{rectification}. We replace the canny edge detector by running a gaussian blur operation on the image and then apply an adaptive threshold. The output is then passed to the probabilistic hough line transform to extract line segments. We made this choice because we intend to apply RANSAC on gridlines in the gram region which is structurally different from edges between different objects (Figure \ref{fig:fig6}).

The output of probabilistic hough line transform is organized into a set of line segments $\{l_i\}$. We define the model $M(l_1,l_2)$ be the representation of a vanishing point at the intersection of $l_1,l_2$, Given a model $M$, the vote of a particular line segment ${l_i}$ for $M$ is 
$$   vote(l_i,M)=\left\{
\begin{array}{ll}
      \left\Vert  l_i  \right\Vert_{2} & \theta < 5^{\circ} \\
      0 & otherwise \\
\end{array} 
\right. 
$$
where $\theta$ is the intersection angle between $l_i$ and the line passing through the vanishing point represented by $M$ and the center of the line segment $l_i$. We apply a standard RANSAC using this metric by randomly generating model proposals and picking the one with the highest total votes. After one vanishing point $M_1$ is identified, all inliners $\{l_i | vote(l_i,M_1)>0\}$ are removed and we reapply RANSAC on the remaining line segments to get the second vanishing point. With two vanishing points, we compute an inverse perspective transform and finally rectify the image (Figure \ref{fig:figrec}).

\subsubsection{Mask-RCNN-Based Approach}
\begin{figure}
	\centering\
	   \includegraphics[width=0.5\linewidth]{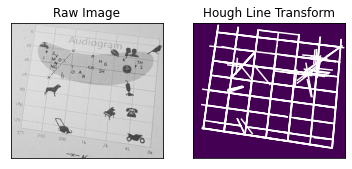}

% 	\fbox{\rule[-.5cm]{4cm}{4cm} \rule[-.5cm]{4cm}{0cm}}
	\caption{An example of interference caused by shapes in the gram region. When the raw image contains irregular shapes in the gram region, the hough line transform yields lines that are not gridlines.}
	\label{fig:figobs}
\end{figure}

Since the previous approach heavily relies on line grids, it is susceptible to interference from other patterns and shapes in the gram region. In some variations of audiograms (not included in our dataset), there are many congested shapes in the gram region which affects the reliability of hough line transform (Figure \ref{fig:figobs}). To address this issue, we proposed an alternative border-based approach utilizing Mask-RCNN.

\begin{figure}
	\centering\
	   \includegraphics[width=0.8\linewidth]{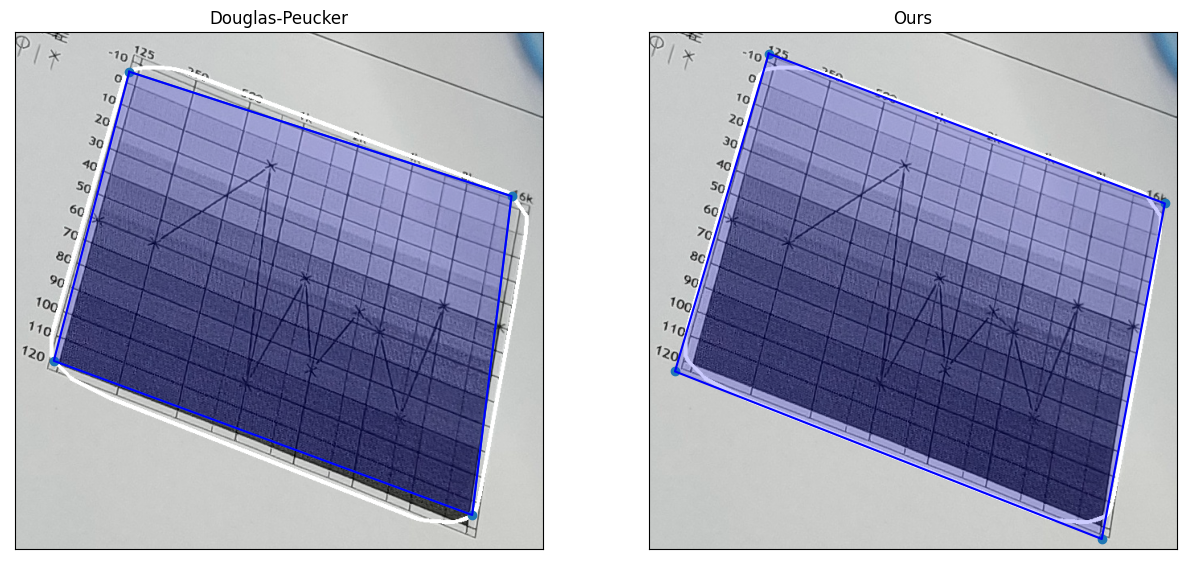}

% 	\fbox{\rule[-.5cm]{4cm}{4cm} \rule[-.5cm]{4cm}{0cm}}
	\caption{A comparison between Douglas-Peucker and our approach. The white lines are the bounding polygons predicted by Mask-RCNN. The blue lines are the approximate quadrilaterals generated by two algorithms. Douglas-Peucker yield suboptimal results because its output is a subset of vertices of the original polygon, which contains "rounded edges". }
	\label{fig:figapprox}
\end{figure}

Mask-RCNN is a variant of RCNN modified for image segmentation tasks \cite{he2017mask}. In addition to bounding boxes, Mask-RCNN is able to generate bounding polygons of detected objects from predicted masks. We define the chart region as the box containing line grids. It is different from the gram region in that it does not contain axis labels. Hence, it has separate annotations from gram regions in our dataset. We use a Mask-RCNN to extract the bounding polygon of the chart region, and approximate the bounding polygon with a quadrilateral. While similar polygon simplification tasks are often achieved by the Douglas-Peucker algorithm \cite{visvalingam1990douglas},  it is not directly applicable as the mask polygons predicted by our network tend to have "rounded edges." Since the four corners are not vertices of the mask polygon, Douglas-Peucker will yield sub-optimal results as its output is a subset of the original polygon's vertices. We propose an alternative greedy algorithm to address this issue. We intend to find four straight lines and make their intersections four corners of the quadrilateral. We first apply Douglas-Peucker with a very small threshold ($0.005\times perimeter$) such that the long edges will be smoothed but the "rounded corners" will not be reduced. To find the first line, we find the longest edge in the smoothed polygon and extend it to a line. We remove all edges in close proximity to this line and repeat the same process to find the second, third and fourth line. With four lines we get six intersections (they can be at infinity when lines are strictly parallel). We remove the furthest 2 points and use the remaining four to construct the approximate quadrilateral of the original polygon. This method can find corners that are not a vertex of the original polygon and hence it is more desirable than applying Douglas-Peucker naively (Figure \ref{fig:figapprox}).

We use the four points of the fitted quadrilateral as estimations of four corners of the chart region. From these four points, we construct an inverse perspective transform that would rectify these four points to four vertices of a square.

\subsection{Axis and Mark Detection}
With the image rectified, we can proceed to detect marks and axis. This task is no more than a typical multi-class object detection task. We apply the same Faster-RCNN as in 4.1 with the only difference being that we have 24 classes (two types of marks for left ear and right ear, 14 hearing level ticks, and 8 frequency ticks) instead of one.

After we extract the bounding boxes of axis labels and marks, we calculate the centroid of each bounding box and consider them to be the coordinates of each mark/axis label. Since the two axes have no overlapping ticks, we can separate the axis labels into two groups purely based on their numerical value. We apply RANSAC linear regression for each group and fit two lines representing the direction of two axes. RANSAC is introduced to improve the resilience against some misclassified instances.

\begin{figure}
	\centering\
	   \includegraphics[width=0.5\linewidth]{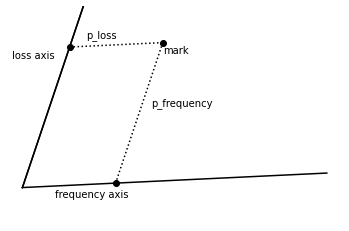}

% 	\fbox{\rule[-.5cm]{4cm}{4cm} \rule[-.5cm]{4cm}{0cm}}
	\caption{Projection of marks. $p_{frequency}$ and $p_{hl}$ are the magnitude of projected vectors on respective axis. }
	\label{fig:figproj}
\end{figure}

For each mark and axis, we calculate their projections onto the fitted axes and acquire $(p_{frequency},p_{hl})$, where $p_{hl}$ is the distance between the origin and the projected points on the hearing level axis and $p_{frequency}$ is the distance between the origin and the projected points on the frequency axis (Figure \ref{fig:figproj}). 

We want to construct a a mapping $g:\mathbb{R}^2\rightarrow \Gamma$ that maps a  $(p_{frequency},p_{hl})$ tuple to an element in $\Gamma$ (the finite set of all possible mark coordinates as described in 2.1). We fist construct two mappings $g_f,g_l: \mathbb{R} \rightarrow \mathbb{R} $ such that $g_f$ maps $p_{frequency}$ to a real valued frequency and $g_l$ maps $p_{hl}$ to a real valued hearing level. These two mappings are individually fitted by RANSAC linear regression on the $p_{hl},p_{frequency}$ values of respective axis labels and their numerical values in dB HL and Hz. $g_l$ is directly fitted since the hearing level ticks are linear. For $g_f$, we take the log of frequency value before passing it to a RANSAC regressor as the frequency ticks are in log scale.

\begin{figure}
	\centering\
	   \includegraphics[width=0.5\linewidth]{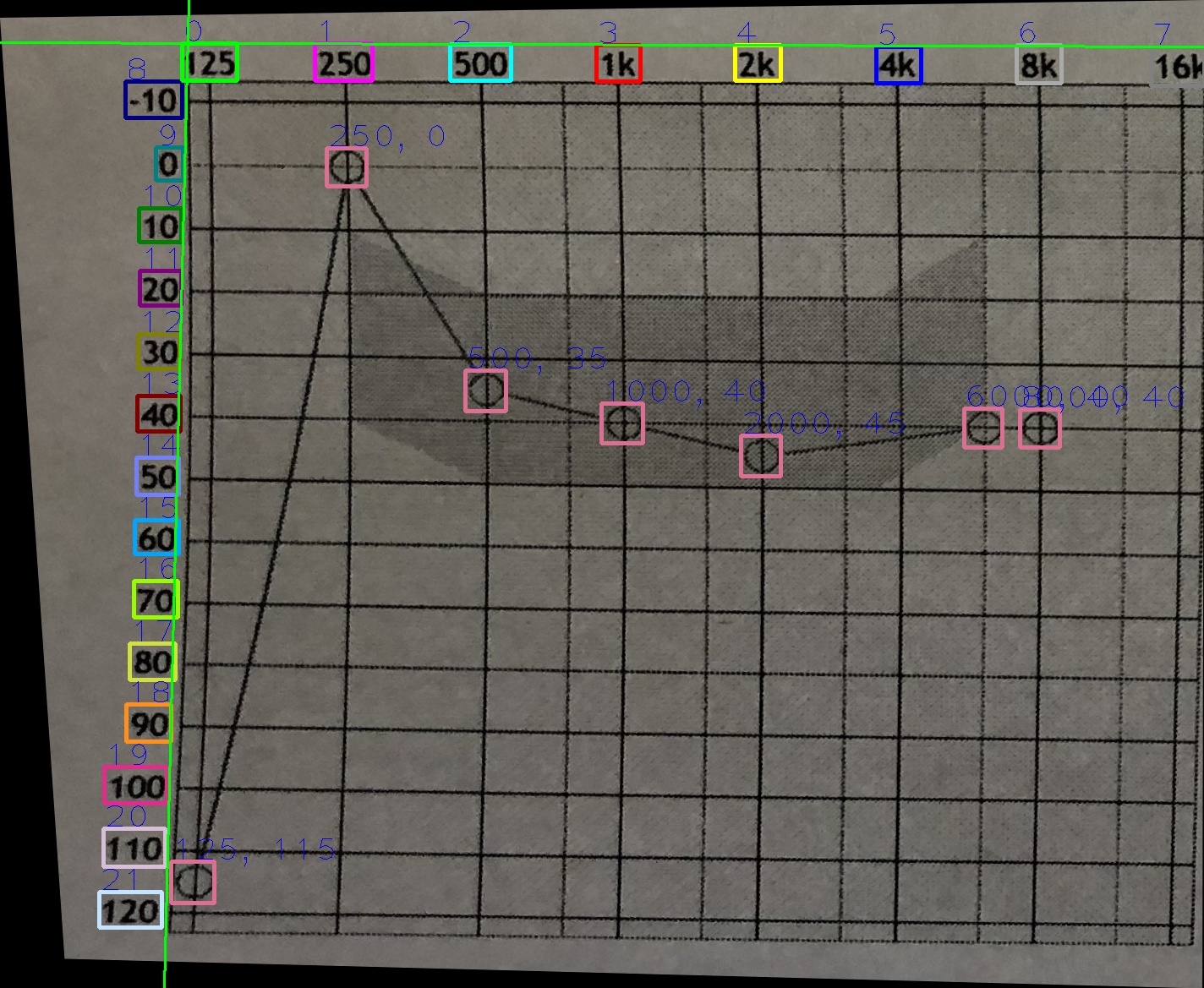}

% 	\fbox{\rule[-.5cm]{4cm}{4cm} \rule[-.5cm]{4cm}{0cm}}
	\caption{An example of MAIN's final outputs. The lines are two fitted axis. The rectangles are bounding boxes of axis labels and marks. }
	\label{fig:fig12}
\end{figure}

With these two mappings, we construct $g:\mathbb{R}^2\rightarrow \Gamma$ such that $g(p_{frequency},p_{hl})$ equals to the closet neighbor of $(g_f(p_{frequency}),g_l(p_{hl}))$ in $\Gamma$. For example, if $(g_f(p_{frequency}),g_l(p_{hl}))$ is $(130,14)$, it will be mapped to $(125,15)$. An example of model output is shown in Figure \ref{fig:fig12}.

\section{Experiments}
\subsection{Model Implementation and Data Preparation}
We implemented Faster-RCNN and Mask-RCNN using PyTorch and detectron2 library \cite{wu2019detectron2}. We implemented our line-detection-based perspective rectification algorithms by modifying Chilamkurthy's work \cite{rectification}. We implemented our Mask-RCNN based algorithms from scratch utilizing various publicly available libraries such as OpenCV. All RANSAC processes in the paper are implemented by the scikit-learn library.

We randomly split the 420 raw images into a training set of 352 images and a test set of 68 images. All images have gram (level 2) annotations, and therefore these two sets are also the training and test set for the gram detector. We have 410 cropped images with annotations of bounding polygons of chart regions (level 3) and they are split into a training set of 369 images and a test set of 41 images. We have 223 cropped images with annotations of axis and marks (level 4) and they are split into a training set of 203 images and a test set of 20 images. It should be noted that while the split is not identical, no cropped images in any of the training set comes from raw images in the test set, which is used in our final evaluation.
\subsection{Training}
All the models are trained on one Nvidia GPU. The Faster-RCNNs were trained for 100 epochs and the Mask-RCNN was trained for 20 epochs. The learning rate of Faster-RCNN used for gram detection is $10^{-5}$ while the learning rate of Faster-RCNN used for axis and label detection is $10^{-4}$. The learning rate of Mask-RCNN is $10^{-4}$. We used the \textit{faster\_rcnn\_X\_101\_32x8d\_FPN\_3x} pre-trained weights from detectron2 library to initialize our Faster-RCNN and \textit{mask\_rcnn\_R\_50\_FPN\_3x} to initialize Mask-RCNN \cite{wu2019detectron2}.  Results show that all the networks achieved a satisfying performance on the test set on their respective tasks (Table \ref{table:table4}).

\subsection{Evaluation}
We evaluate our multi-stage model on the test set of 68 camera photos. We compared the performance of two proposed perspective rectification methods. Results show that both methods are able to improve the model performance on camera photos. We also test our model on a set of 30 scanned images for comparison. Since there is no perspective distortion, we did not apply perspective rectification. The results show that our model can handle scanned images without ever being exposed to such images during the training. In fact, it performances significantly better on scanned images (Table \ref{table:table3}). These results show that our proposed baseline is feasible. To the best of our knowledge, there are no comparable prior works. Hence we did not compare our baseline with other works.

% \usepackage{booktabs}
\begin{table}
\centering\
\caption{Test Set Performance}
\begin{tabular}{@{}ccccl@{}}
                                     & \multicolumn{3}{c}{Photo}                       & Scanned                               \\ \midrule
                                     & No Rectification & Line Detection & Mask RCNN   & \multicolumn{1}{c}{No Rectification} \\ \midrule
Label Total Accuracy                 & 0.765, 0.727$^*$     & 0.853, 0.834    & 0.843, 0.842 & 0.987, 0.991             \\
Frequency Accuracy                   & 0.967, 0.918      & 0.977, 0.955    & 0.963, 0.962 & 0.987, 0.991                          \\
HL Accuracy                        & 0.809, 0.768      & 0.878, 0.858    & 0.868, 0.867 & 0.987, 0.991                          \\
Label Total Accuracy $\pm 5$$^{**}$ & 0.937, 0.890      & 0.960, 0.938    & 0.950,0.949 & 0.987.0.991                          \\ \bottomrule
\end{tabular}

\vspace{1ex}
\raggedright\
\begin{itemize}
  \item[*] The numbers are $recall, precision$ respectively, where recall is defined as (number of correct predictions)/ (number of ground truth marks) and precision is defined as (number of correct predictions)/ (number of predicted marks)
  
  \item[**] This metric assumes all predictions that are off by at most 5dB HL in hearing level as correct.
\end{itemize}
\label{table:table3}

\end{table}

\begin{table}
\centering\
\caption{Test Set Performance of Different Stages}
\begin{tabular}{cccc}
                            & AP    & AP50  & AP75  \\ \hline
Gram (Faster-RCNN)           & 0.907 & 0.987 & 0.987 \\
Axis and Marks (Faster-RCNN) & 0.621 & 0.850 & 0.822 \\
Chart Region (Mask-RCNN)     & 0.965 & 0.965 & 0.965 \\ \hline
\end{tabular}
\label{table:table4}
\end{table}

\section{Conclusion}

There is a growing demand to automatize the process of audiogram interpretation, and our work shows that it is feasible to reconstruct audiogram information from camera photos and scanned images by leveraging state-of-art neural networks.  Our baseline has an accuracy of 98\% on scanned images which suggests that it can be an effective and robust replacement of manual interpretation for audiologists with access to a scanner. An 84\% accuracy on camera photos suggests that such architecture can also be applied in scenarios where users need to take photos of their audiograms using a smartphone, which can be useful in a pandemic lockdown, the process of autonomous fitting of OTC hearing-aids, or in areas with limited access to audiologists.

Experimental results suggest that one major challenge in audiogram interpretation is the distortions caused by varying lighting conditions and camera angles. They also showed that our proposed rectification algorithms can mitigate the impact of perspective distortion. 

To the best of our knowledge, we are the first to establish a benchmark dataset and propose a baseline model for audiogram interpretation. In future works, we will try to adjust our model so that it can be applied in circumstances where the hearing levels of the left and right ears are represented in a single audiogram by two lines with different marks. We may also try to differentiate audiograms of air conduction and bone conduction audiometry which differ in their notations.

\bibliographystyle{unsrt}
\bibliography{references}  
\end{document}